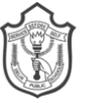

The Ethics of Robotics

Kush Agrawal, Delhi Public School, R.K Puram
2010

Kush Agrawal

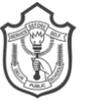

## Abstract

"1. A robot may not injure a human being or, through inaction, allow a human being to come to harm.
2. A robot must obey any orders given to it by human beings, except where such orders would conflict with the First Law.
3. A robot must protect its own existence as long as such protection does not conflict with the First or Second Law." [16]                     - Isaac Asimov

These three laws first appeared together in Isaac Asimov's story 'Runaround'[15] after being mentioned in some form or the other in previous works by Asimov. These three laws commonly known as the three laws of robotics are the earliest forms of depiction for the needs of ethics in Robotics. In simplistic language Isaac Asimov is able to explain what rules a robot must confine itself to in order to maintain societal sanctity. However, even though they are outdated they still represent some of our innate fears which are beginning to resurface in present day 21st Century. Our society is on the advent of a new revolution; a revolution led by advances in Computer Science, Artificial Intelligence & Nanotechnology. Some of our advances have been so phenomenal that we surpassed what was predicted by the Moore's law. With these advancements comes the fear that our future may be at the mercy of these androids. Humans today are scared that we, ourselves, might create something which we cannot control. We may end up creating something which can not only learn much faster than anyone of us can, but also evolve faster than what the theory of evolution has allowed us to. The greatest fear is not only that we might lose our jobs to these intelligent beings, but that these beings might end up replacing us at the top of the cycle. The public hysteria has been heightened more so by a number of cultural works which depict annihilation of the human race by robots. Right from Frankenstein[14] to I, Robot[14] mass media has also depicted such issues.

This paper is an effort to understand the need for ethics in Robotics or simply termed as Roboethics. This is achieved by the study of artificial beings and the thought being put behind them. By the end of the paper, however, it is concluded that there isn't a need for ethical robots but more so ever a need for ethical roboticists.

# Introduction

The term 'Robot' first appeared in the play R.U.R (Rossum's Universal Robots) by Czech writer Karel Čapek in the year 1921[6][25]. The term arises from the Polish word 'Robota' which literally means self-labourer, or hard worker[13]. Although the term Robot can cover a multitude of definitions; there is general consensus that a Robot is a programmable device/machine which has been created to imitate the actions of an intelligent creature (usually a human being)[20]. It has to be able to receive input from its surroundings which may be in any of the forms of stimuli. Also, it must display corporeal skills such as movement and manipulation of its environment.

Over the past 50 years, Robots have weaved their way into our Society. So much so, that some of our work has become dependent on their proper functioning. Robots such as the Roomba have even gone on to become household names. The last few decades has bought a radical shift in our reaction towards robots and the field in general. Humans have become more accepting, and ironically at the same time more sceptical. However, this field is one of the key components of our future, and this is the right time to discuss and avoid any problems that we may face in the future. Shying away from Robotics is nothing less than shying away from the future.

# Superiority

There are many capabilities that robots have, and we lack. For instance we can receive 4 forms of input from environmental stimulus: touch, smell, sight and sound. A robot on the other hand can mimic what we achieve using a variety of sensors ranging from infrared sensors to ultrasonic sound wave sensors. Their senses are only limited by our imagination, and as the years go by, Scientists all over the world are coming up with new techniques to give input to machines be it visual, auditory or plain old text. Another, aspect in which Robots do hold an advantage over us is their work ethic. They would be any production line owners delight since getting tired or bored is out of the question. Their superiority is also apparent in the fact that they can carry out complex calculations in a matter of seconds. The Tianhe-I, presently the worlds faster super computer, can carry out 2½ quadrillion floating point operations per second (FLOPS)[26]. Computer chips today run a million times faster than the speed of a human neuron, and this speed is not likely to reduce by any means in the future.

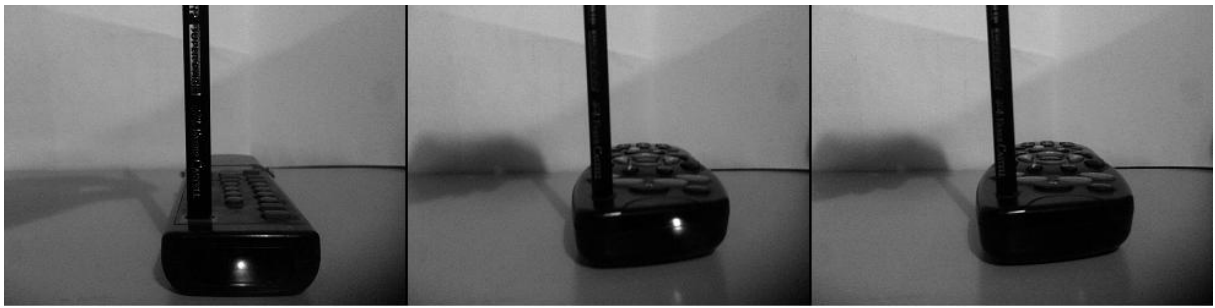

(A simple example of how a robots vision may be augmented. The two images on the left represent the fact that cameras are Infra-Red sensitive, however infra-red light lies outside the scope of the vision of Human beings, and as a result what a human being may see is depicted by the image on the right)

## Robots=Slaves?

The question naturally arises "What role would robots fit into in our future?" Will they, as their literal Polish sense means, act as slaves to the human race?[6][25] Or will they be given equal rights as human beings? This is one of the most complex questions that we face today. And the answer entirely depends on the types and kinds of robots that we build. Sentience will be the deciding factor with regard to the identity that these machines will hold in our society. Sentience is one's ability to perceive and feel. In essence Sentience can be described as one's consciousness of one's own existence. The irony is that most artificial machines, and in fact all our machines at present are not sentient. If that is the case, then there is no logic in giving robots an autonomous status. However, if we do become successful in constructing sentient beings, it would be an interesting circumstance to provide these beings with a charter of rights of their own. The recognition of sentient beings, apart from us, will be an extremely important and necessary step in our future, since not providing them with a status will hinder their free will and as a result affect their sentience.

## Are Robots morally responsible?

At some level Robots are a reflection of our own desires and aspirations and at many instances they can be used to judge the society in which they were created[1]. As was said above, Robots carry out instructions exactly as they were inputted to them (sometimes a bit too literally), so who is at fault in case they commit a mistake, or worse a crime? The answer again lies in the question whether they are sentient beings and whether they are awarded autonomous status. Once a Robot has been declared as free willed, it may be held accountable for its own actions since such status will only be awarded as long as a Robot can discern independently right from wrong, correct from incorrect. So if a free willed robot commits a crime, even though it may be under the influence of someone else, it may still be held partially accountable for the crime. There still do arise a variety of questions on the moral accountability of Robots such as:

- "Who's responsible for the undesirable actions of a robot?"
- "Who decides what types and kinds of robots to be made?"
- "Who regulates what they monitor and the data processed by them?" and most importantly
- "Who governs the robots and their actions?"

These questions will have to be answered in the near future, and the earlier that they are answered the better it will be for us (human kind i.e.).

## Limits

Now, the natural question that arises is that should there be a restriction on the types and kinds of Robots that we as humans can design and develop? Or should it be a field where one and all are free to express and create whatever breed of robots that they wish to. Robotics, as such, is a very sensitive field since Robots are closer to Humans than Computers or any other machines that we might have ever created both morphologically and literally[9][10]. The simple reason for that is their shape and form. They remind us of ourselves. This very simple fact changes everything since it means that at some level they represent us. The best approach in this regard will be that of Utilitarianism so that Robots can work towards the greater good of a greater number of people. A set of rules must be put into place so that developmental resources are directed towards projects that meet the required standards of Utilitarianism. Therefore work on Robots that fall into the following parameters must be dissuaded:

- Robots that injure innocent life forms including but not restricted to Human beings
- Robots that process or record sensitive, restricted information in an unauthorised manner
- Robots that propagate/promote war, or any unhealthy form of human interaction
- Robots that cause harm to the environment for the benefit of a few people
- Robots that may deviate from their predefined objectives

However, Research & Development on the Robots which fall into the following parameters should be promoted:

- Robots that prevent injury/harm to other life forms.
- Robots that help conserve/ uphold our environmental laws or in any other forms promote environmental welfare.
- Robots that promote healthy livelihood and good living habits.

## Reality Warp

If we do end up creating sentient beings, which can rival us in most aspects, how will we discern between who's a Robot, and who's a human being? As the field of Robotics progresses further, the thin line that lies between us and robots will start to blur out and eventually disappear[10]. More and more Human beings will receive Bionic implants that augment their abilities, and even more number of Robots will be created that match humans in most aspects. How then do we plan to differentiate between humans and Robots? Or do we need to differentiate at all?  The answer probably is that we do. It's a simple fact that we are human beings and they are our creations. Even though we may give them an independent status it is important to identify between artificial and non-artificial since separate laws may hold true for them. But the more important questions is "How do we differentiate?" One may answer that Creativity is the answer, or that Robots cannot be creative and hence that could be the next TURING test[11]. But the fact is that as more advances are made, we may be able to produce artificial life that is creative as well or at least it may be able to fake some sort of creative ability[11]. The only way we can ensure that we do not get fooled is to implement some sort of design feature that will easily help us spot a robot out of a crowd. For instance a simple test for the effectiveness of the design feature could be to ask someone to spot a Robot out of a group which includes humans, and then compare the efficiency of the results.  This warp in reality, can not only affect the societal sanctity, but also the sanity of beings[11]. One needs to be aware of their reality, whether they are a human or a humanoid, otherwise it may lead to a range of emotional and psychological disorders related to identity loss and doubt on one's own identity.

Kush Agrawal

## Loss of work

"Laziness may appear attractive, but work gives satisfaction."

-Anne Frank

As Robotic systems increase in complexity, there can and will be an increase in loss of jobs. More and more workers will be replaced by Robots because of the:

- Cheaper long terms costs associated with Robotic machines
- More efficiency- Robots can cover more mechanical ground in a shorter time span
- No liabilities in terms of insurance, employment rights etc.

However, it is imperative that Humans do find jobs since lack of them will lead to gloom, despair and a global economic crisis since the number of dependants will increase. There are a few solutions to these problems. People must be retrained to take up jobs, in which they cannot be replaced by Robots, for instance someone is required to repair the Robot in case of a failure[12]. Future laws must also be altered to take into consideration the changing role that Robots play in society. Laws must be put in place to protect the interests of mankind such as a cap on the percentage of the Robotic force at a company. Another advantage that we as Humans posses over Robots is that time is on our side. We will only be able to build robots that are advanced enough to replace Painters, Musicians and Artists after we have gained enough knowledge of the subject, and have gained mastery over the intricacies and subtleties of it. This phenomenon can attributed to the Moravec's paradox devised by Artificial Intelligence researchers Hans Moravec, Marvin Minsky & Rodney Brooks in the 1980's. As Steven Pinker writes "The main lesson of thirty-five years of AI research is that the hard problems are easy and the easy problems are hard. The mental abilities of a four-year-old that we take for granted – recognizing a face, lifting a pencil, walking across a room, answering a question – in fact solve some of the hardest engineering problems ever conceived.... As the new generation of intelligent devices appears, it will be the stock analysts and petrochemical engineers and parole board members who are in danger of being replaced by machines. The gardeners, receptionists, and cooks are secure in their jobs for decades to come."[27]

## Uses

The role that Robots will come to play in the future of our Society is entirely dependent on how we decide to use them. At this present instance, though we are way beyond the point of no return[12], we may still be able define how Robots become a part of our society in the future. It is imperative that utilization of Robots is capped in some fields, yet promoted in the others. Creating Robots that take care of household chores may be a double edged sword. On one hand we may expect that the Robots will help reduce the work load on the Human beings and will allow them to concentrate on other more important activities, while on the other it may turn humans into lazy couch potatoes. Humans will become obese, lazy and frustrated for the lack of work. There may also be instances when Humans may depend on Robots to carry out tasks such as baby sitting or teaching. Such utilization can and will have dire consequences for it will hinder the growth and development of the child. There was even a recent study which proved that monkeys which grew under Robotic care, turned out to be social outcasts, unable to mate and interact with the other monkeys[18]. Handing

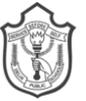

Kush Agrawal 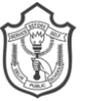

over our jobs to Robots might have consequences as well, since it will do away with creative and original thought. Another instance where Robotics could be harmful for our future is in Sports and Athletics, as the market for both sports and robotics grows side by side, there will come a time when both of them will begin to diffuse into each other. Athletes and Sports persons will consider augmentation of their playing abilities by the use of bionic implants. To ensure that a level playing field is maintained, rules must also be put in place to ensure that Cyborgs and Humans do not compete on the same field unknowingly. Robots however do and should play a vital role in the future of medicine and surgery since they eliminate many unwanted factors such as contamination.

## Discussion & Conclusion

Mankind is on the advent of a revolution led by the field of Robotics. As is with any revolution there must be some set of ground rules established to ensure that this technological revolution does not catch us off guard or overwhelm us. As we see above, there is a need for ethics in the field of Robotics. But the natural question arises whether we can build Robots complex enough to govern themselves. The fact remains that such a future is far off, and we need regulations to ensure that our community is not harmed by the misuse/negligent use of these beautiful machines. Robots are powerful; they can do our taxes and at the same time create economic havoc; they can build cars and at the same time destroy them; they can stimulate us, enchant us & elevate us. But the fact remains that they must do so in a controlled manner, and to ensure that, there must be some room for ethics in Robotics. How do we do that? Creating ethical Robots is equivalent in complexity to creating sentient robots, so it is not ethical Robots that we seek. It is ethical Roboticists that we ask for. For it is only then that these exciting machines may take the shape and form that we want them to.

Kush Agrawal

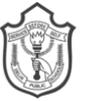

# Citations & References